\documentclass{ecai}
\usepackage{graphicx}
\usepackage{latexsym}
\usepackage{amsmath}
\usepackage{amssymb}
\usepackage{multirow}

\begin{document}

\begin{frontmatter}

\title{Building a Few-Shot Cross-Domain Multilingual NLU Model for Customer
Care}

 \author[A]{\fnms{Saurabh}~\snm{Kumar}\orcid{0009-0006-5396-2723}\thanks{Corresponding Author. Email: saurabh.kumar1@walmart.com}}
 \author[B]{\fnms{Sourav}~\snm{Bansal}\orcid{0009-0004-5447-7186}\thanks{Work was done as an intern at Walmart Global Tech, Bangalore, India}}
 \author[A]{\fnms{Neeraj}~\snm{Agrawal}}\orcid{0000-0003-1496-6618} 
 \author[A]{\fnms{Priyanka}~\snm{Bhatt}}\orcid{0009-0009-6431-0490} 

\address[A]{Walmart Global Tech, Bangalore, India}
\address[B]{Indian Institute of Technology, Delhi}

\begin{abstract}

Customer care is an essential pillar of the e-commerce shopping experience with companies spending millions of dollars each year, employing automation and human agents, across geographies (like US, Canada, Mexico, Chile), channels (like Chat, Interactive Voice Response (IVR)), and languages (like English, Spanish). SOTA pre-trained models like multilingual-BERT, fine-tuned on annotated data have shown good performance in downstream tasks relevant to Customer Care. However, model performance is largely subject to the availability of sufficient annotated domain-specific data. Cross-domain availability of data remains a bottleneck, thus building an intent classifier that generalizes across domains (defined by channel, geography, and language) with only a few annotations, is of great practical value.

In this paper, we propose an embedder-cum-classifier model architecture which extends state-of-the-art domain-specific models to other domains with only a few labeled samples. We adopt a supervised fine-tuning approach with isotropic regularizers to train a domain-specific sentence embedder and a multilingual knowledge distillation strategy to generalize this embedder across multiple domains. The trained embedder, further augmented with a simple linear classifier can be deployed for new domains. Experiments on Canada and Mexico e-commerce Customer Care dataset with few-shot intent detection show an increase in accuracy by 20-23\% against the existing state-of-the-art pre-trained models.

\end{abstract}

\end{frontmatter}

\begin{figure*}
\centering
    \includegraphics[width=0.9\textwidth]{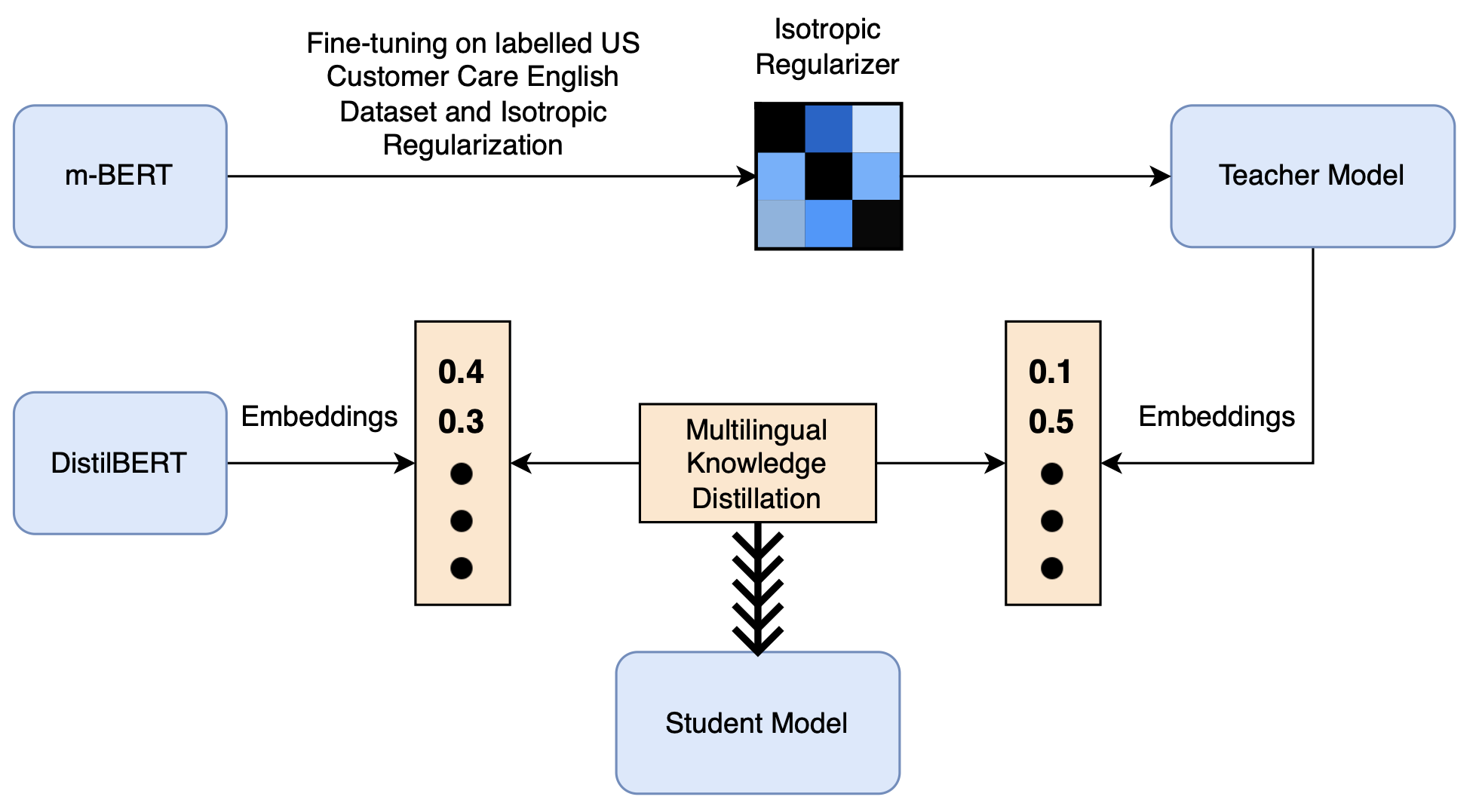}
    \caption{Proposed Model Training Methodology. Multilingual BERT is fine-tuned and regularized to produce a teacher model. Distilled multilingual BERT is trained using knowledge distillation to generate the student model.}
    \label{fig:model}
\end{figure*}

\section{Introduction}
Task-oriented dialogue systems have been widely deployed for a variety of sectors ranging from Shopping \cite{10.5555/3298023.3298238} to Customer Care \cite{cs_dialog_sys}, to provide an interactive experience. Intent classification is a core module that drives the working of task-oriented dialogue systems. Thus, building accurate classifiers is important for the development of such systems.

Traditional methods have fine-tuned pre-trained language models for various NLP (natural language processing) applications and have achieved state-of-the-art performance. While fine-tuning certainly generates improvements, a significant challenge is the availability of sufficient annotated data as the performance is highly dependent on it. However, large annotated datasets are usually available only for selected languages and tasks. This poses a challenge in the development of models which could serve users in multiple domains (defined by channel, geography, and language) when only a few labeled samples are available. Hence, building a generalized few-shot cross-domain classifier that leverages information across domains is an important problem.

Recently, the problem has received considerable attention. Some recent works have used generation-based methods \cite{xu2022generating}, induction network \cite{geng-etal-2019-induction} and metric learning \cite{nguyen-etal-2020-dynamic} for building few-shot classifiers. The focus of these works has been designing new algorithms which consequently come with complicated models. With rising interest and credibility of large-scale pre-trained language models, many works have fine-tuned these models \cite{zhang-etal-2020-discriminative, zhang-etal-2021-effectiveness-pre} which have yielded promising results in various downstream tasks. However, the construction of large annotated datasets is an expensive and laborious task. Consequently, the focus of many recent works has shifted to unsupervised pre-training of language models \cite{gururangan-etal-2020-dont, gu-etal-2021-pral} with specifically designed objectives to suit downstream tasks. Such methods have generated good results but utilize large-scale corpora of unsupervised data with closely related semantics.

An important prerequisite for models to work across multiple domains and languages is the alignment of vector spaces across domains. This enables the model to produce language-agnostic sentence representations which can capture rich semantic information, highly useful for downstream classification tasks. In this paper, we present a method to train a generalized sentence embedding model useful for cross-domain classification tasks. The solution provides the capability to expose domain-specific state-of-the-art models to new domains with just a few labelled utterances.

The proposed method utilizes a knowledge distillation strategy \cite{DBLP:journals/corr/abs-2004-09813} in order to extend the intelligence of the existing domain-specific model, called the teacher model to a cross-domain multilingual model, called the student model. The teacher model is a fine-tuned large language model, which is used to generate sentence embeddings of the utterances for the source domain. The student model is trained to mimic the teacher model in a multilingual setup, that is, it maps utterances with similar meaning in other languages close to the original utterance. The trained student model can now be deployed in new domains by training a classifier with just a few examples.

The proposed method also adopts isotropic regularizers for improving sentence representations generated by the models. \cite{zhang2022finetuning} mentions that fine-tuned PLMs (Pre-Trained Language models) may suffer from anisotropy, which could be the reason for the sub-optimal performance of PLMs on downstream tasks. We utilize a correlation matrix-based regularizer to regularize the supervised training of the teacher model. This improves the embeddings generated by the teacher model, which in turn leads to a more accurate student model.

Extensive experiments were conducted to validate the performance of the model. Experiments on Canada e-commerce Customer Care domain English dataset with few-shot intent detection have shown an increase in accuracy by 23\% against the existing state-of-the-art pre-trained models. Experiments across Mexico e-commerce Customer Care domain Spanish Dataset have also shown a boost in accuracy of 20\% against the existing state-of-the-art models.

\begin{figure*}
\centering
    \includegraphics[width=0.9\textwidth]{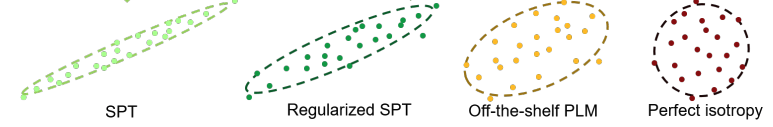}
    \caption{Change in isotropy of embedding space due to regularization during supervised pre-training. SPT denotes supervised pre-training (fine-tuning an off-the-shelf PLM on a set of labeled utterances), which makes the feature space more anisotropic}
    \label{fig:isotrop}
\end{figure*}
The main contributions of our paper are summarised as follows:

\begin{itemize}
    \item We propose an embedder-cum-classifier model architecture that extends domain-specific monolingual models to other domains with only a few labelled utterances.
    \item We fine-tune pre-trained language models using labelled data, introduce isotropic regularizers to improve sentence representations in source domains, and transfer the intelligence to a student model using a knowledge distillation process in a multilingual setup, which allows it to function in multiple domains.
    \item Experiments demonstrate that the proposed model achieves an improvement of 23\% and 20\% in Canada and Mexico Customer Care Dataset respectively against the existing state-of-the-art pre-trained models.
\end{itemize}

The rest of the paper is organized as follows. Section \ref{section:methodology} presents the methodology used for training. Section \ref{section:experiments} presents the results of experiments used for evaluating the model.

\begin{figure*}
\centering
    \includegraphics[width=1\textwidth]{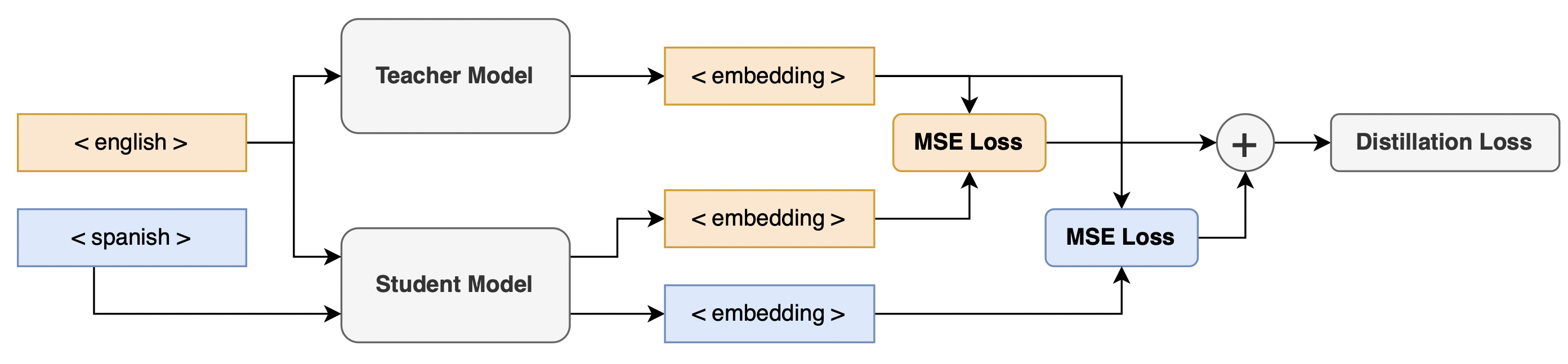}
    \caption{Given parallel translated sentences across languages (e.g., English and Spanish), the teacher model generates embeddings for source language sentences and the student model generates embeddings for both source and target language sentences. The MSE loss between embeddings of the student and teacher model is added to create distillation loss.}
    \label{fig:kd}
\end{figure*}

\section{Methodology}
\label{section:methodology}
This section describes the proposed model in detail. Figures \ref{fig:model} and \ref{fig:kd} show the overall architecture of the model.

\subsection{Text Encoder}
\label{le}
Multilingual BERT \cite{devlin-etal-2019-bert}, one of the state-of-the-art encoders for multilingual texts, has been used as the text encoder. BERT uses WordPiece algorithm to tokenize the input text. Given an input text $x$, it is tokenized into sequence of tokens $x_1, x_2,... x_{n-2}$. BERT adds [CLS] and [SEP], two special tokens indicating the beginning and the end of the sequence. Therefore, the final sequence of tokens of length $n$ is represented as:

$$
    x = \{\textnormal{[CLS]}, x_1, x_2, ..., x_{n-2}, \textnormal{[SEP]}\} 
    \label{eq1}
$$


BERT encodes input tokens and outputs encodings corresponding to each token. We use encoding corresponding to [CLS] token as the representation of the sentence fed.

$$
    h = \textnormal{BERT}(x)
$$

Where, $h$ $\in \mathbb{R}^{d}$, $d$ is size of the sentence embedding generated.

\subsection{ Supervised Pre-training }
Recent works have shown that further pre-training off-the-shelf PLMs using dialogue corpora \cite{henderson-etal-2019-training, peng-etal-2020-shot} is beneficial for task-oriented downstream tasks such as intent detection.

Similar to \cite{zhang-etal-2021-effectiveness-pre, zhang2022finetuning}, our pre-training method relies on the existence of a labeled dataset $\mathcal{D}_{\text {source }}^{\text {labeled }}=\left\{\left(x_{i}, y_{i}\right)\right\}$, where $y_{i}$ is the label for utterance $x_{i}$. 

  Given $\mathcal{D}_{\text {source }}^{\text {labeled }}=\left\{\left(x_{i}, y_{i}\right)\right\}$ with $N$ different classes, we employ a simple method to fine-tune BERT. Specifically, a linear layer is attached on top of BERT as the classifier, i.e.,

$$
p\left(y \mid h_{i}\right)=\operatorname{softmax}\left(\mathbf{W} h_{i}+\mathbf{b}\right) \in \mathbb{R}^{N}
$$

where $h_{i} \in \mathbb{R}^{d}$ is the feature representation of $x_{i}$ given by the $[C L S]$ token, $\mathbf{W} \in \mathbb{R}^{N \times d}$ and $\mathbf{b} \in \mathbb{R}^{N}$ are parameters of the linear layer. The model parameters $\theta=\{\phi, \mathbf{W}, \mathbf{b}\}$, with $\phi$ being the parameters of BERT, are trained on $\mathcal{D}_{\text {source }}^{\text {labeled }}$ with a cross-entropy loss:

$$
\theta^{*}=\underset{\theta}{\arg \min } \mathcal{L}_{c e}\left(\mathcal{D}_{\text {source }}^{\text {labled }} ; \theta\right)
$$

\begin{table*}[]
   \centering
    \begin{tabular}{c|c|c|c}
    \hline
         Dataset & Language  & \#intent & \#utterances \\
         \hline
         
         US Customer Care Chat Dataset & English & 26 & 37,322\\
         Mexico Customer Care Chat Dataset & Spanish & 16 & 3,677\\
         Canada Customer Care Chat Dataset & English & 21 & 1,025\\
         US Customer Care IVR Dataset & English & 14 & 5,994\\

        \hline
         
    \end{tabular}
    \caption{Statistics of datasets used for experimentation.}
    \label{tab:datasetdetails}
\end{table*}
\begin{table*}[]
\resizebox{\textwidth}{!}{%
   \centering
    \begin{tabular}{c c  c c c c c c c c c c c}
    
        \noalign{\vskip 0.04in} 
         \hline
         \noalign{\vskip 0.04in} 
          Method & \multicolumn{4}{c}{\textbf{Mexico CC Domain Dataset} (Spanish)} &  \multicolumn{4}{c}{\textbf{Canada CC Domain  Dataset} (English) }  &  \multicolumn{4}{c}{\textbf{US CC IVR Domain Dataset} (English) }  \\
        
          & 2-shot & 5-shot & 10-shot & 20-shot &2-shot & 5-shot & 10-shot & 20-shot &2-shot & 5-shot & 10-shot & 20-shot\\
         
         \noalign{\vskip 0.04 in} 
         \hline
         \noalign{\vskip 0.04in} 
         
         \noalign{\vskip 0.04in} 
         \textbf{Pre-Trained Models}\\
         \noalign{\vskip 0.04in} 
         \hline
         \noalign{\vskip 0.04in}
         distilled multilingual BERT & 9.4 & 13.4 & 20.9 & 29.9 &15.7 & 22.1 & 24.0 & 30.3 & 11.1 & 21.1 & 23.5 & 30.4\\
         multilingual BERT & 10.3 & 13.6 & 21.6 & 30.1  &    10.2  &  20.1 &23.2 & 31.9& 9.7 & 19.2 & 22.5 & 29.8\\
         LaBSE  & 25.3 & 39.5 & 51.1 & 57.5 &    27.6  & 38.2 & 46.4 & 52.8& 41.1 & 63.2 & 65.1 & 73.2\\
         use-cmlm-multilingual & 33.9 & 50.8 & 53.2 & 56.9 & 23.2 & 33.1 & 35.03 & 46.9 & 40.7 & 61.8 & 64.6 & 72.3\\
         \noalign{\vskip 0.04in} 
         \hline
         \noalign{\vskip 0.04in} 
         \textbf{Pre-Trained Knowledge-Distilled Models}\\
         \noalign{\vskip 0.04in} 
         \hline
         \noalign{\vskip 0.04in} 

         distiluse-base-multilingual-cased-v1 & 22.7 & 52.6 & 50.3 & 58.6 & 33.1 &48.4  & 54.0 & 56.7 & 48.7 & 65.6 & 69.4 & 74.6 \\
         distiluse-base-multilingual-cased-v2 & 18.1 & 45.2 & 53.9 & 55.7 & 32.7  & 48.8 & 46.1 &54.7& 47.4 & 65.2 & 68.1 & 74.4\\
         paraphrase-multilingual-MiniLM-L12-v2 & 14.9 & 54.7 & 58.0 & 58.6 & 29.1 &44.9 & 47.2 & 51.6 & 57.4 & 67.6 & 67.3 & 73.4\\
         paraphrase-multilingual-mpnet-base-v2 & 16.0 & 57.4 & 59.7 & 63.1 & 34.3  & 56.3 & 61.0 & 61.4 & 58.1 & 68.1 & 68.3 & 74.2\\
         
         \noalign{\vskip 0.04 in} 
         \hline
         \noalign{\vskip 0.04 in} 
         \textbf{Our Models}\\
         \noalign{\vskip 0.04 in} 
         \hline
         \noalign{\vskip 0.04 in} 
         TEACHERv1 (Fine-tuned BERT Model)  & 54.8 & 57.7 & 58.8 & 60.1 & 68.1  & 68.9 & 67.3 & 67.7 & 55.9 & 68.7 & 70.1 & 76.1\\
         TEACHERv2 (Fine-tuned BERT Model with Cor-Reg) & 55.3 & 58.7 & 59.9 & 61.2 & 69.4  & \textbf{70.1} & \textbf{68.2} & 68.8 & 56.0  & 68.9 & 72.1 & 77.0\\
         STUDENTv1 (Knowledge Distilled from Teacher V1) & 69.5 & 76.9 & 77.9 & 78.5 & 68.5  & 69.3 & 67.8 & 69.7 & 58.1 & 68.3 & 69.6 & 75.4\\
         STUDENTv2 (Knowledge Distilled from Teacher V2) & \textbf{70.1} & \textbf{77.4} & \textbf{78.6} & \textbf{79.1} & \textbf{69.7}  & \textbf{70.0} & \textbf{68.2} & \textbf{69.7} & \textbf{58.6}  & \textbf{69.2} & \textbf{70.4} & \textbf{76.5}\\

         \hline

    \end{tabular}}
    \caption{N-Shot experimental results comparing the performance of the proposed model with other state-of-the-art models on Mexico CC Domain (Spanish Language), Canada CC Domain (English Language) datasets and US CC IVR Domain Dataset (English).}
    \label{tab:results}
\end{table*}
\subsubsection{Fine-tuning leads to Anisotropy}
Recent work from \cite{zhang2022finetuning} mentions that further pre-training off-the-shelf PLMs may lead to anisotropy, which has been identified as the reason for the sub-optimal performance of PLMs on downstream tasks. Hence, we utilize isotropic regularizers as mentioned in the next section.

\subsection{Regularizing Supervised Pre-training with Isotropization}
\label{subsection:3.3}
Isotropization techniques can be applied to adjust the embedding space and yield significant performance improvement in many tasks. Figure \ref{fig:isotrop} illustrates the effect of supervised pre-training and regularised supervised pre-training on isotropy. To mitigate the anisotropy of the PLM fine-tuned by supervised pre-training, as proposed by \cite{zhang2022finetuning}, we use a joint training objective by adding a regularization term $\mathcal{L}_{\text {reg }}$ for isotropization:

$$
\mathcal{L}=\mathcal{L}_{\text {ce }}\left(\mathcal{D}_{\text {source }} ; \theta\right)+\lambda \mathcal{L}_{\text {reg }}\left(\mathcal{D}_{\text {source }} ; \theta\right)
$$

where $\lambda$ is a weight parameter. The aim is to learn intent detection skills while maintaining an appropriate degree of isotropy. We use a Correlation-matrix-based regularizer \cite{zhang2022finetuning}

$$
\mathcal{L}_{\mathrm{reg}}=\|\boldsymbol{\Sigma}-\mathbf{I}\|
$$

where $\|\cdot\|$ denotes Frobenius norm, $\mathbf{I} \in \mathbb{R}^{d \times d}$ is the identity matrix, $\boldsymbol{\Sigma} \in \mathbb{R}^{d \times d}$ is the correlation matrix with $\boldsymbol{\Sigma}_{i j}$ being the Pearson correlation coefficient between the $i_{\text {th }}$ dimension and the $j_{\text {th }}$ dimension. $\boldsymbol{\Sigma}$ is estimated with utterances in the current batch. By pushing the correlation matrix towards the identity matrix during training, we can learn a more isotropic feature space.

\subsection{Knowledge distillation }
With regularized supervised pre-training as explained in section \ref{subsection:3.3}, we already have a model which can generate accurate embeddings for a domain. We use this model as the teacher model ($M$) and transfer the intelligence to the student model ($\hat{M}$). The multilingual knowledge distillation process has been explained in Figure \ref{fig:kd}. The teacher model maps the sentences in the source domain to a high-dimensional vector space. 

For the multilingual distillation process, we utilize an unsupervised dataset of parallel translated sentences, denoted as $D = \{((s_1, t_1), \dots, (s_n, t_n))\}$, where $s_i$ is the sentence in the source domain language and $t_i$ is the sentence in the target domain language. The training process for the student model minimizes the mean-squared loss between embeddings generated by the teacher and student model. The mean-squared loss is taken between the embeddings of the teacher model in source language and the embeddings of the student model in source language as well as the embeddings of the teacher model in source language and embeddings of the student model in target language. The exact objective for a batch $\beta$ is mentioned in the equation below.
$$
\frac{1}{|\beta|}\sum\limits_{j \in \beta} [(M(s_j) - \hat{M}(s_j))^2 + (M(s_j) - \hat{M}(t_j))^2]
$$

The loss function tries to remove the language bias, that is, sentences with similar meanings but in different languages are mapped closer than sentences in the same language with different meanings. In our experiments, we use the multilingual BERT as the teacher model $M$ and multilingual DistilBERT as the student model $\hat{M}$.

 \subsection{Few-shot Intent Classification}
The student model ($\hat{M}$) generated by Knowledge Distillation can be immediately used as a feature extractor for novel few-shot cross-domain multilingual intent classification tasks when used along with a classifier on top of it. The classifier can be a parametric one such as Support Vector Machine (SVM) or a non-parametric one such as nearest neighbor. A parametric classifier will be trained with the few labeled examples provided in a task and make predictions on the unlabeled queries. As evaluated in the experiments presented in the next section, a simple linear classifier suffices to achieve very good performance, owing to the effective utterance representations produced by our Knowledge Distilled Student Model.

\section{Experiments}
\label{section:experiments}
In order to assess the effectiveness of the proposed method, we run several experiments across different datasets as mentioned below.

\subsection{ Experiment Setup}
\textbf{Datasets and Evaluation Metrics} To perform supervised pre-training of the teacher model, we use a labeled US Customer Care English Chat dataset which consists of manually labeled issue summary texts written by users interacting with the e-commerce chatbot. For multilingual knowledge-distilled training of the student model, we use an unsupervised subset of the same dataset and its Spanish translation. For evaluation of cross-domain performance, we use Canada E-commerce Customer Care domain dataset which consists of utterances in the English language, Mexico E-commerce Customer Care Domain Chat dataset which consists of utterances in the Spanish language and US Customer Care IVR dataset which consists of utterances in the English language. Dataset statistics are summarized in Table \ref{tab:datasetdetails}.

\begin{figure*}
\centering
    \includegraphics[width=1\textwidth]{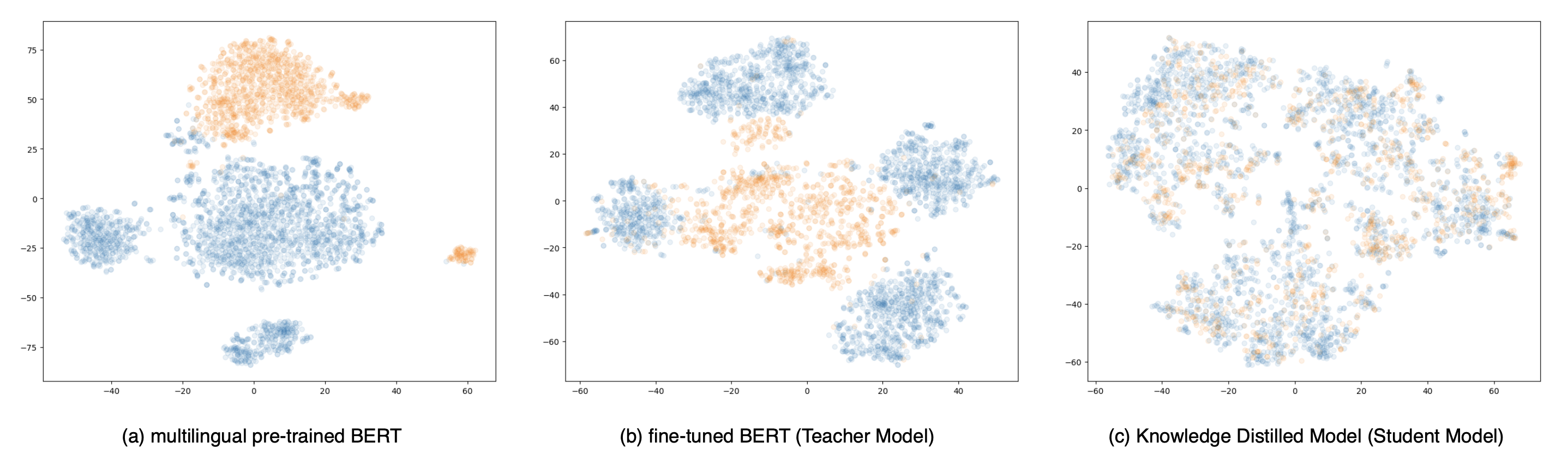}
    \caption{Embedding space visualizations plotted in two-dimensional space using t-SNE. The blue dots represent embeddings for English language sentences and the orange dots represent embeddings for Spanish language sentences. Sentences were randomly selected from the top four most occurring intents for visualization purposes.}
    \label{fig:visual-1}
\end{figure*}

\textbf{Implementation Details} We conduct experiments on two of the most popular PLMs for multilingual texts, BERT \cite{devlin-etal-2019-bert} Multilingual version and DistilBERT \cite{sanh2020distilbert} Multilingual version. BERT Multilingual has been used as the teacher model and DistilBERT Multilingual as the student model. Both models are trained using the process outlined in the previous section. As explained in section 2.1, for few-shot intent classification tasks, the embedding of [CLS] from the student model is used as the utterance representation. We employ Support Vector Machine (SVM) as the classifier.

\textbf{Baseline} 
We compare our model to the following strong baselines. Firstly, for pre-trained large language models, we present a comparison against DistilBERT multilingual \cite{devlin-etal-2019-bert}, BERT multilingual \cite{sanh2020distilbert}, LaBSE \cite{DBLP:journals/corr/abs-2007-01852} and use-cmlm-multilingual \cite{DBLP:journals/corr/abs-1810-04805}. LaBSE is a language-agnostic BERT embedder supporting 109 languages and use-cmlm-multilingual is model trained using Conditional Masked Language Modelling for improved sentence embeddings. Further, we compare our proposed method against four multilingual knowledge-distilled models pre-trained by authors of \cite{DBLP:journals/corr/abs-2004-09813} and made available in the Sentence Transformers library. distiluse-base-multilingual-cased-v1 and distiluse-base-multilingual-cased-v2 are multilingual distilBERT models with teacher model as USE, supporting 15 and 50 languages respectively. paraphrase-multilingual-MiniLM-L12-v2 is trained with the teacher model as paraphrase-MiniLM-L12-v2 and the student model as Multilingual-MiniLM-L12-H384 while paraphrase-multilingual-mpnet-base-v2 is trained with teacher model as paraphrase-mpnet-base-v2 and the student model as xlm-roberta-base.

\textbf{Training details}  The complete code has been written using Python and PyTorch library. We use Hugging Face implementation of \textit{bert-base-multilingual-cased} and \textit{distilbert-base-multilingual-cased}. Adam \cite{kingma2017adam}, with a learning rate of $2\mathrm{e}{-5}$ has been used as the optimizer. The model is trained with Nvidia V100 GPU.

\subsection{ Experiment Results}
Table \ref{tab:results} compares the performance of the proposed approach against the baselines. The following observations can be made. All our models consistently outperform all the baselines by a significant margin, indicating the robustness of the proposed methods against cross-domain multilingual data. For the 5-shot classification task, our Student Model ($\hat{M}$) outperforms the strongest baseline paraphrase-multilingual-mpnet-base-v2 by an absolute margin of 20\% on Mexico e-commerce Customer Care Domain Spanish Language Dataset and 14\% on Canada e-commerce Customer Care Domain English Language Dataset. The gain is attributed to supervised isotropic fine-tuning and knowledge-distillation strategy. In the IVR Domain, the student model shows a marginal improvement of 1\%. This could be attributed to the very different nature of IVR Data which has been converted to text using Automatic Speech Recognition (ASR) and usually comes with smaller and broken phrases.

On the Spanish language dataset of Mexico e-commerce Customer Care Domain, both student models outperform the corresponding teacher models by a large margin of around 10\%. This is because the teacher models are fine-tuned on the English language dataset of US e-commerce Customer Care Domain, hence, unable to learn embeddings in the multilingual space. While for the English language dataset of Canada e-commerce Customer Care domain and IVR Domain, the student models and the corresponding teacher models yield similar performance since the utterances are in the same language as the US Customer Care Domain dataset (used for training). This also explains why the model achieves maximum performance on 5-shot training for Canada e-commerce Customer Care domain, as the model seems to overfit on adding more samples.

\subsection{ Analysis and Ablation Study}

\subsubsection{ Effect of Isotropy}
Table \ref{tab:results} shows that introducing a Correlation Matrix based regularizer for Isotropization leads to around 1\% improvement in the accuracy score. We adjust the weight parameter $\lambda$  of Cor-Reg to analyze the relationship between the isotropy of the feature space and the performance of few-shot intent detection. Similar to \cite{zhang2022finetuning}, we observed that moderate isotropy is helpful. The performance of few-shot intent detection task increases with an increase in isotropy only up to a certain extent and decreases after that. This could be attributed to the fact that high isotropization may reduce the essence of supervised fine-tuning. Therefore, it is crucial to find a good balance between learning intent detection skills and learning anisotropic feature space.

\subsubsection{Effect of Knowledge Distillation}
We use t-SNE for visualization of high-dimensional embeddings in two-dimensional space as shown in Figure \ref{fig:visual-1}. Figure \ref{fig:visual-1}(a) shows embeddings generated by the pretrained multilingual BERT which exhibit very weak alignment among English and Spanish Language sentence representations. Though the graph shows some clustering for utterances of different intents, the clusters for English and Spanish languages are far apart. Figure \ref{fig:visual-1}(b) depicts embeddings for BERT fine-tuned on English labeled dataset (TEACHERv2). The clusters for English sentence embeddings representing different intents have improved, however, representations for Spanish have degraded and are clustered in the center far apart from their English counterparts. Figure \ref{fig:visual-1}(c) shows the representation generated by the student model (STUDENTv2). It shows that the representation of sentences from English and Spanish Language are close. Further, there are no clusters which offer a generalized embedder that can be used for different classification tasks. Clearly, knowledge distillation helps produce language-agnostic representations, that is, representations are neutral with respect to the language.

\section{Conclusion}

 In this paper, we proposed a training method that utilizes a multilingual knowledge distillation strategy in order to extend a domain-specific state-of-the-art model to a cross-domain multilingual model. The paper also discusses the use of isotropic regularizers based on correlation matrix in order to improve sentence representations generated by the models. Combination of these two approaches leads to significantly improved performance on cross-domain multilingual few-shot intent detection task. Experiments confirm the feasibility and practicality of developing a few-shot intent classifier that could be deployed in multiple domains even with different languages, saving the cost involved in data annotation. Detailed ablation studies prove the effectiveness of the different components.


\bibliography{ecai}

\begin{thebibliography}{10}

\bibitem{DBLP:journals/corr/abs-1810-04805}
Jacob Devlin, Ming{-}Wei Chang, Kenton Lee, and Kristina Toutanova, `{BERT:} pre-training of deep bidirectional transformers for language understanding', {\em CoRR}, {\bf abs/1810.04805}, (2018).

\bibitem{devlin-etal-2019-bert}
Jacob Devlin, Ming-Wei Chang, Kenton Lee, and Kristina Toutanova, `{BERT}: Pre-training of deep bidirectional transformers for language understanding', in {\em Proceedings of the 2019 Conference of the North {A}merican Chapter of the Association for Computational Linguistics: Human Language Technologies, Volume 1 (Long and Short Papers)}, pp. 4171--4186, Minneapolis, Minnesota, (June 2019). Association for Computational Linguistics.

\bibitem{DBLP:journals/corr/abs-2007-01852}
Fangxiaoyu Feng, Yinfei Yang, Daniel Cer, Naveen Arivazhagan, and Wei Wang, `Language-agnostic {BERT} sentence embedding', {\em CoRR}, {\bf abs/2007.01852}, (2020).

\bibitem{geng-etal-2019-induction}
Ruiying Geng, Binhua Li, Yongbin Li, Xiaodan Zhu, Ping Jian, and Jian Sun, `Induction networks for few-shot text classification', in {\em Proceedings of the 2019 Conference on Empirical Methods in Natural Language Processing and the 9th International Joint Conference on Natural Language Processing (EMNLP-IJCNLP)}, pp. 3904--3913, Hong Kong, China, (November 2019). Association for Computational Linguistics.

\bibitem{gu-etal-2021-pral}
Jing Gu, Qingyang Wu, Chongruo Wu, Weiyan Shi, and Zhou Yu, `{PRAL}: A tailored pre-training model for task-oriented dialog generation', in {\em Proceedings of the 59th Annual Meeting of the Association for Computational Linguistics and the 11th International Joint Conference on Natural Language Processing (Volume 2: Short Papers)}, pp. 305--313, Online, (August 2021). Association for Computational Linguistics.

\bibitem{gururangan-etal-2020-dont}
Suchin Gururangan, Ana Marasovi{\'c}, Swabha Swayamdipta, Kyle Lo, Iz~Beltagy, Doug Downey, and Noah~A. Smith, `Don{'}t stop pretraining: Adapt language models to domains and tasks', in {\em Proceedings of the 58th Annual Meeting of the Association for Computational Linguistics}, pp. 8342--8360, Online, (July 2020). Association for Computational Linguistics.

\bibitem{henderson-etal-2019-training}
Matthew Henderson, Ivan Vuli{\'c}, Daniela Gerz, I{\~n}igo Casanueva, Pawe{\l} Budzianowski, Sam Coope, Georgios Spithourakis, Tsung-Hsien Wen, Nikola Mrk{\v{s}}i{\'c}, and Pei-Hao Su, `Training neural response selection for task-oriented dialogue systems', in {\em Proceedings of the 57th Annual Meeting of the Association for Computational Linguistics}, pp. 5392--5404, Florence, Italy, (July 2019). Association for Computational Linguistics.

\bibitem{kingma2017adam}
Diederik~P. Kingma and Jimmy Ba.
\newblock Adam: A method for stochastic optimization, 2017.

\bibitem{nguyen-etal-2020-dynamic}
Hoang Nguyen, Chenwei Zhang, Congying Xia, and Philip Yu, `Dynamic semantic matching and aggregation network for few-shot intent detection', in {\em Findings of the Association for Computational Linguistics: EMNLP 2020}, pp. 1209--1218, Online, (November 2020). Association for Computational Linguistics.

\bibitem{peng-etal-2020-shot}
Baolin Peng, Chenguang Zhu, Chunyuan Li, Xiujun Li, Jinchao Li, Michael Zeng, and Jianfeng Gao, `Few-shot natural language generation for task-oriented dialog', in {\em Findings of the Association for Computational Linguistics: EMNLP 2020}, pp. 172--182, Online, (November 2020). Association for Computational Linguistics.

\bibitem{DBLP:journals/corr/abs-2004-09813}
Nils Reimers and Iryna Gurevych, `Making monolingual sentence embeddings multilingual using knowledge distillation', {\em CoRR}, {\bf abs/2004.09813}, (2020).

\bibitem{sanh2020distilbert}
Victor Sanh, Lysandre Debut, Julien Chaumond, and Thomas Wolf.
\newblock Distilbert, a distilled version of bert: smaller, faster, cheaper and lighter, 2020.

\bibitem{cs_dialog_sys}
Manex Serras, Naiara Perez, M.~Torres, Arantza Pozo, and Raquel Justo, `Topic classifier for customer service dialog systems', pp. 140--148, (09 2015).

\bibitem{xu2022generating}
Jingyi Xu and Hieu Le.
\newblock Generating representative samples for few-shot classification, 2022.

\bibitem{10.5555/3298023.3298238}
Zhao Yan, Nan Duan, Peng Chen, Ming Zhou, Jianshe Zhou, and Zhoujun Li, `Building task-oriented dialogue systems for online shopping', in {\em Proceedings of the Thirty-First AAAI Conference on Artificial Intelligence}, AAAI'17, p. 4618–4625. AAAI Press, (2017).

\bibitem{zhang2022finetuning}
Haode Zhang, Haowen Liang, Yuwei Zhang, Liming Zhan, Xiao-Ming Wu, Xiaolei Lu, and Albert Y.~S. Lam.
\newblock Fine-tuning pre-trained language models for few-shot intent detection: Supervised pre-training and isotropization, 2022.

\bibitem{zhang-etal-2021-effectiveness-pre}
Haode Zhang, Yuwei Zhang, Li-Ming Zhan, Jiaxin Chen, Guangyuan Shi, Xiao-Ming Wu, and Albert~Y.S. Lam, `Effectiveness of pre-training for few-shot intent classification', in {\em Findings of the Association for Computational Linguistics: EMNLP 2021}, pp. 1114--1120, Punta Cana, Dominican Republic, (November 2021). Association for Computational Linguistics.

\bibitem{zhang-etal-2020-discriminative}
Jianguo Zhang, Kazuma Hashimoto, Wenhao Liu, Chien-Sheng Wu, Yao Wan, Philip Yu, Richard Socher, and Caiming Xiong, `Discriminative nearest neighbor few-shot intent detection by transferring natural language inference', in {\em Proceedings of the 2020 Conference on Empirical Methods in Natural Language Processing (EMNLP)}, pp. 5064--5082, Online, (November 2020). Association for Computational Linguistics.

\end{thebibliography}
\end{document}